\documentclass[letterpaper, 10 pt, conference]{ieeeconf}  

\IEEEoverridecommandlockouts                              

\usepackage{graphics} 
\usepackage{amsmath} 
\usepackage{amssymb}  
\usepackage{subfiles}
\usepackage{makecell}
\usepackage{graphicx}
\usepackage{multirow}
\usepackage{soul}
\usepackage{changes}
\usepackage{tikz}
\usepackage{xcolor}
\usepackage{wasysym}
\usepackage[utf8]{inputenc}

\newcommand*\circled[1]{\tikz[baseline=(char.base)]{
            \node[shape=circle,draw,inner sep=1pt]  (char)  { \scriptsize #1};}}

\usepackage{hyperref}
\usepackage{cleveref}
\usepackage[T1]{fontenc}
\usepackage{caption}

\crefformat{section}{\S#2#1#3} 
\crefformat{subsection}{\S#2#1#3}
\crefformat{subsubsection}{\S#2#1#3}

\title{
\parbox{\textwidth}{ {\vspace{-30mm} \small Accepted to the 2025 IEEE International Conference on Robotics and Automation (ICRA). This is a preprint version.}} \\ 
\Large \bf \vspace{-6mm}
Hierarchically Accelerated Coverage Path Planning for Redundant Manipulators 
\vspace{-4mm}
}

\author{Yeping Wang and Michael Gleicher
\thanks{ Both authors are with the Department of Computer Sciences, University of Wisconsin-Madison, Madison, WI 53706, USA $\hspace*{0.8in}$
{\tt\small [yeping|gleicher]@cs.wisc.edu}}%
\thanks{This work was supported by Los Alamos National Laboratory, the Department of Energy, and National Science Foundation Award 2007436.}
}

\makeatletter
\let\@oldmaketitle\@maketitle
\renewcommand{\@maketitle}{\@oldmaketitle
   \vspace{-1mm}
    \includegraphics[width=7in]{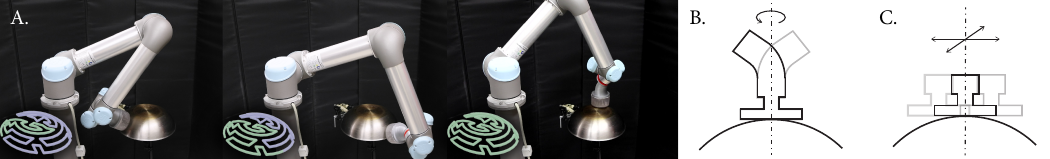}
    \captionof{figure}{We present an effective and efficient coverage path planning approach that exploits a robot manipulator's redundancy and task tolerances to minimize joint space costs. (A) A robotic arm performs a mock wok polishing task, with the coverage path shown in the lower left. This task has (B) rotational redundancy around the tool's principal axis and (C) translational tolerance tangential to the wok surface, as the finishing disk can have multiple contact points with the wok.  Due to the redundancy, infinite possible motions can cover the surface, and our approach finds one that minimizes joint space costs. }
    \label{fig: teaser}
    }
\newcommand{\algorithmfootnote}[2][\footnotesize]{%
  \let\old@algocf@finish\@algocf@finish
  \def\@algocf@finish{\old@algocf@finish
    \leavevmode\rlap{\begin{minipage}{\linewidth}
    #1#2
    \end{minipage}}%
  }%
}
\makeatother

\begin{document}
 \vspace{-10mm}

\maketitle
\thispagestyle{empty}
\pagestyle{empty}

\addtocounter{figure}{-1} 

\begin{abstract}
Many robotic applications, such as sanding, polishing, wiping and sensor scanning, require a manipulator to dexterously cover a surface using its end-effector. In this paper, we provide an efficient and effective coverage path planning approach that leverages a manipulator's redundancy and task tolerances to minimize costs in joint space. We formulate the problem as a Generalized Traveling Salesman Problem and hierarchically streamline the graph size. Our strategy is to identify guide paths that roughly cover the surface and accelerate the computation by solving a sequence of smaller problems.
We demonstrate the effectiveness of our method through a simulation experiment and an illustrative demonstration using a physical robot.
\end{abstract}

\vspace{-2mm}
\section{Introduction}
\vspace{-1mm}
In applications such as sanding, polishing, wiping, or sensor scanning, a robot manipulator needs to move its end-effector to cover an entire surface. In cases where the robot has redundant degrees of freedom or the application allows certain tolerances, the system obtains some \textit{flexibility}, \textit{i.e.}, there are infinitely many inverse kinematics (IK) solutions for positioning the end-effector at each surface point. A coverage path planner should exploit the flexibility to find a joint motion that fully covers the surface, while minimizing objectives like joint movements and maintaining short computation times.



This paper provides a solution to the problem of coverage path planning for redundant robot manipulator systems, aiming to minimize specific costs in joint space.
Following prior work \cite{hess2012null, kaljaca2020coverage, zhang2024jpmdp}, we formulate this problem as a Generalized Traveling Salesman Problem (GTSP), where the objective is to find the shortest route in a graph that visits exactly one node from each predefined set. Each node represents an IK solution, with sets grouping IK solutions that cover the same surface point, and edge weights representing the movement costs between two IK configurations. A GTSP route corresponds to a sequence of IK solutions that move the end-effector to cover all required surface points. Using this formulation, prior work \cite{hess2012null, kaljaca2020coverage, zhang2024jpmdp} randomly samples multiple IK solutions for each surface point, constructs a graph, and solves the GTSP. 
However, this approach often leads to a large graph, and due to the NP-hard nature of GTSP, it requires significant computation time \cite{hess2012null, suarez2018robotsp}. 

In this paper, we present an efficient and effective coverage path planning approach for redundant manipulator systems. 
In contrast to prior work which randomly samples IK solutions, we incorporate a strategy to identify IK samples that are likely to lead to good solutions. By identifying guide paths that roughly cover the surface and sample IK solutions near them, our strategy accelerates the computation by solving a sequence of smaller GTSP. Our approach enables frequent identification of effective solutions within a significantly reduced timeframe. We provide an open-source implementation of our approach\footnote{Open-sourced code \url{https://github.com/uwgraphics/arm_coverage}}.

The central contribution of this paper is an effective and efficient coverage path planning approach for robot manipulators (\cref{sec:technical_details}). To facilitate understanding of the problem, we describe and analyze two baseline approaches based on prior work (\ref{sec:cart-tsp} and \cref{sec:joint_gtsp}). All three approaches were evaluated using four simulated benchmarks (\cref{sec:evaluation}). Our results show that the proposed approach generates higher-quality motions with shorter computation time compared to the baselines.

\vspace{-1mm}
\section{Related Work}
\vspace{-1mm}


\vspace{-1mm}
\subsection{Coverage Path Planning for Robot Manipulators}
\vspace{-1mm}
Coverage path planning is required for a robot manipulator to perform tasks such as polishing \cite{schneyer2023segmentation}, bush trimming \cite{kaljaca2020coverage}, surface cleaning \cite{hess2012null, sakata2023coverage, moura2018automation}, surface inspection \cite{jing2017sampling, jing2018computational, leidner2016robotic}, surface disinfection \cite{zhang2024jpmdp, thakar2022area}, and post-processing after 3D printing \cite{do2023geometry}. Coverage path planning approaches for robotic arms generally fall into two categories. The first group focuses on planning Cartesian paths for specific tasks, which the robotic arm then follows. These approaches often involve specialized end-effectors tailored for specific tasks, such as sanding using different parts of the sanding disk \cite{schneyer2023segmentation}, post-processing after 3D printing using a blast nozzle  \cite{do2023geometry},  spraying with aerosol cans \cite{thakar2022area}, wiping with a combination of three actions \cite{leidner2016robotic}, and cleaning surfaces of unknown geometry \cite{moura2018automation}. 
These works assume that the robot can accurately move the end-effector along the planned path. Planning in the Cartesian space, they can not achieve objectives in joint space such as minimizing joint movements \cite{zhang2024jpmdp}. 

The second group of methods plans directly in joint space for both non-redundant \cite{yang2024improved, yang2020cellular, yang2020non} and redundant robot manipulators \cite{hess2012null, kaljaca2020coverage, zhang2024jpmdp}. For redundant systems, as outlined in the Introduction, one approach samples multiple IK solutions for each surface point, constructs a joint space graph, and solves a GTSP. However, due to GTSP's NP-hard nature, this method, which we call Joint-GTSP, is sensitive to graph size. To reduce computation time, prior works often restrict the number of IK samplings $m$ for each surface point, \textit{e.g.}, $m{=}16$ \cite{kaljaca2020coverage} and $m{=}9$ \cite{hess2012null}. However, this constraint restricts the exploration of the solution space. 
Due to Joint-GTSP's limitations, prior works \cite{ hess2012null, suarez2018robotsp} use an alternative method that builds a graph in Cartesian space, solves a Traveling Salesman Problem (TSP), and uses a path tracking method to follow the TSP route.  In the Cartesian graph, nodes represent surface points, and edge weights depend on distance and curvature between points. We call this approach Cart-TSP-IKLink, where IKLink is a path tracking method. Although more efficient due to a smaller graph size compared to Joint-GTSP, Cart-TSP-IKLink does not plan in joint space and can not optimize joint-space objectives such as minimizing joint movements. 
We will further describe Joint-GTSP and Cart-TSP-IKLink in \cref{sec:technical_overview} and use them as baselines in our experiments. In this work, we propose a coverage path planning approach that is more efficient and effective than both Joint-GTSP and Cart-TSP-IKLink.

\vspace{-3mm}
\subsection{Generalized Traveling Salesman Problem} \label{sec:gtsp}
\vspace{-1mm}
In addition to coverage path planning for robot manipulators, other robotics problems can be formulated as Generalized Traveling Salesman Problems (GTSP), such as task sequencing for robot manipulators \cite{suarez2018robotsp, saha2006planning, alatartsev2015robotic} and viewpoint coverage path planning for unmanned aerial vehicles \cite{song2018surface, obermeyer2012sampling, dhami2024gatsbi, bahnemann2021revisiting, cao2020hierarchical}.
Both exact and approximate methods are used to solve GTSP \cite{pop2023comprehensive}. While exact methods \cite{pop2007new, noon1991lagrangian, kara2012new} guarantee optimal solutions, they require significant computation time. 
Approximation methods are polynomial time algorithms that
produce approximate solutions. 
Among many approximate GTSP solvers \cite{hu2008effective, helsgaun2015solving, snyder2006random}, 
we choose GLKH \cite{helsgaun2015solving} for its open-source availability \cite{GLKH} and state-of-the-art performance on large scale datasets \cite{suarez2018robotsp, smith2017glns}. 

The solution of GTSP is a \textit{cycle}, \textit{i.e.}, a closed loop with no defined start or end. However, coverage path planning requires a \textit{path}, where the manipulator does not return to the starting point. In order to extract a path from a cycle, Hess \textit{et al.} \cite{hess2012null} select the starting point as the node closest to the robot's current configuration. However, this method does not extract the shortest path even if the cycle is the shortest. In this work, we construct the graph in a way so that a GTSP solution directly yields the shortest path 
 (\cref{sec:joint_gtsp}-Step C).



\vspace{-3mm}
\subsection{Arm Reconfigurations}
\vspace{-1mm}

Most coverage path planners assume that the robot manipulator can cover the given surface as a whole without interruptions. The uninterrupted coverage of some surfaces is infeasible due to the robot’s joint limits, self-collision, or obstacles elsewhere in the environment.
In such cases, it becomes necessary to divide the surface into smaller segments, requiring the robot to perform \textit{arm reconfigurations} -- a process where the robot deviates from the reference surface, repositions itself in configuration space, and resumes task execution. 
Arm reconfigurations, also known as retractions \cite{kaljaca2020coverage} or discontinuities \cite{yang2020cellular}, should be minimized as they increase time and energy consumption.

Previous work has proposed coverage path planners to minimize reconfigurations \cite{kaljaca2020coverage, yang2024improved, yang2020cellular, yang2020non}. 
However, these methods are designed for non-redundant robots and their extension to redundant robots remains unclear. This paper presents a coverage path planner for redundant manipulator systems, with the goal of minimizing reconfigurations.

\vspace{-3mm}
\subsection{Flexibility in Task Execution}
\vspace{-1mm}

In cases where the manipulator is redundant, \textit{i.e.}, the robot has more degrees of freedom than the task requires, or the task has certain tolerances, \textit{i.e.}, allowable position or rotation inaccuracy, there are infinitely many IK solutions that satisfy the task requirements. Such flexibility has been leveraged to achieve active compliance \cite{sadeghian2013task}, emotional expression \cite{claret2017exploiting}, efficient trajectory tracking \cite{wang2025anytime}, and smooth telemanipulation \cite{wang2023exploiting}. In this work, we exploit the flexibility to optimize arm reconfigurations and joint movements in coverage path planning. 

\vspace{-1mm}
\section{Technical Overview} \label{sec:technical_overview}
\begin{figure*} [t]
  \centering
  \includegraphics[width=6.9in]{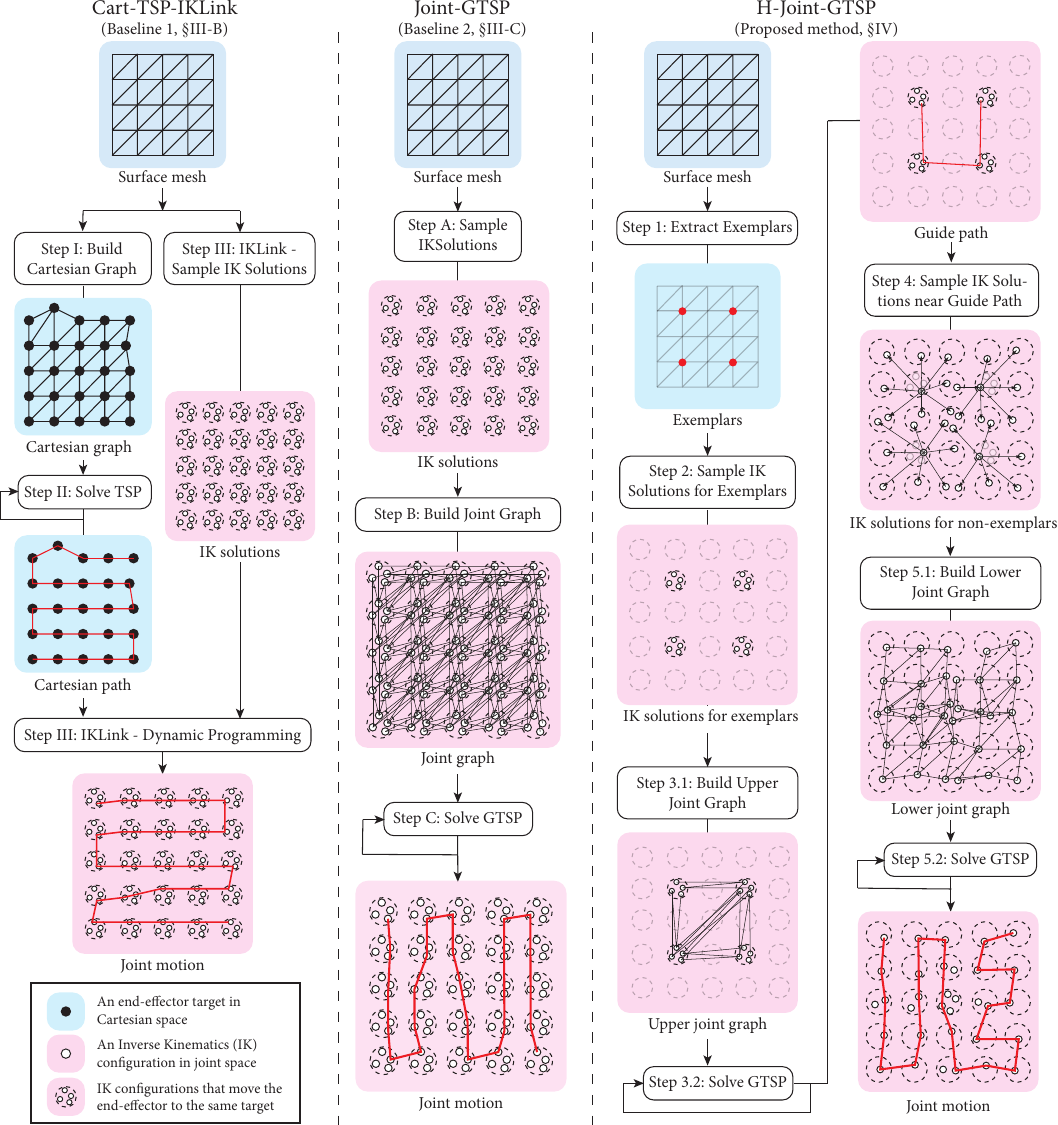}
  \caption{An illustration of the three approaches described in this paper and evaluated in our experiment.}
  \label{fig: approaches}
  \vspace{-7mm}
\end{figure*}

\vspace{-1mm}
In this section, we formalize our problem statement and describe two baseline approaches. Fig. \ref{fig: approaches} provides an illustration of the two baseline approaches and the proposed method. 

\vspace{-2mm}
\subsection{Problem Formulation} 
\vspace{-1mm}


The coverage problem is defined on a surface mesh represented as a connected triangle mesh $\mathcal{M}{=}(V, E)$, where $V{=}\{\mathbf{p}_1, \mathbf{p}_2, ..., \mathbf{p}_n\}$ are the vertices and $E$ represents the edges. 
The number of vertices $n$ defines the level of discretization of the surface. We assume that the vertices are sampled densely enough to accurately represent the shape of the surface and that visiting all vertices is sufficient to cover the surface.
For each vertex $\mathbf{p}_i {\in} \mathbb{R}^3$, we compute its normal vector $\mathbf{n}_i {\in} \mathbb{R}^3$, such that $\lVert \mathbf{n}_i \lVert_2 {=}1$.
We define the \textit{end-effector target} at each vertex as $\mathbf{x}_i{=}(\mathbf{p}_i, \mathbf{n}_i)$, where $\mathbf{x}_i {\in} \mathbb{R}^3 {\times} S^2$. This target provides the position and orientation constraints for the end-effector in 5 degrees of freedom (DoF). 
For a manipulator with $k$ DoF, whose joint configuration is represented by $\mathbf{\theta} {\in} \mathbb{R}^k$, the robot is considered redundant if $k{>}5$.
In addition, the application may allow certain tolerances, \textit{i.e.}, allowable position or rotation inaccuracy.  

The goal of this work is to generate a joint space path $\boldsymbol{\xi} {=} \{\mathbf{\theta}_1, \mathbf{\theta}_2, ..., \mathbf{\theta}_n \}$ that drives the robot's end-effector to visit all end-effector targets on the surface exactly once while minimizing specific costs in joint space. The primary objective is to minimize the number of arm reconfigurations, with the secondary goal of reducing overall joint movements.
We assume that between $\mathbf{\theta}_i$ and $\mathbf{\theta}_{i+1}$ the robot can either linearly move in its joint space or needs to perform an arm reconfiguration using a motion planner.

\vspace{-3mm}
\subsection{Baseline 1: Cart-TSP-IKLink} \label{sec:cart-tsp}
\vspace{-1mm}

We adapt Cart-TSP-IKLinK from prior work \cite{hess2012null, suarez2018robotsp}, incorporating IKLink \cite{wang2024iklink} for the final path following step. This approach constructs a graph in Cartesian space, uses a Traveling Salesman Problem (TSP) solver to find paths that visit each node exactly once, and generates joint motions to track the paths using IKLink. 

\textit{Step I: Construct a Cartesian Graph} --- The Cartesian graph shares the same nodes and edges as the input surface mesh. The edge weights are determined by a combination of the positional and rotational distances.
\vspace{-2mm}
\begin{equation}
    \Delta(\mathbf{x}_i,\mathbf{x}_j) = \lVert \mathbf{p}_i - \mathbf{p}_j \lVert_2 + \alpha \cdot \text{arccos}(\frac{\mathbf{n}_i \cdot \mathbf{n}_j}{\lVert\mathbf{n}_i\lVert_2\lVert\mathbf{n}_j\lVert_2})
    \label{eq:cartesian_dist}
\vspace{-1mm}
\end{equation}
where $\alpha$ is a parameter, set to $\alpha{=}0.1$ in our implementation. 

\textit{Step II: Solve TSP} ---
As discussed in \cref{sec:gtsp}, the solution of TSP is a \textit{cycle}, from which we extract a \textit{path}. We introduce a dummy node to the graph, connected to all other nodes with zero-weight edges, and then convert a TSP solution to a path by removing the dummy node. In this work, we use the LKH \cite{helsgaun2000effective} solver to iteratively find TSP routes, which are then converted into paths.  

\textit{Step III: Track Cartesian Path (using IKLink)} ---
We use IKLink \cite{wang2024iklink} to generate joint movements that move the end-effector along the paths found in Step II. IKLink enables a robot to track Cartesian paths with minimal reconfigurations and joint movements. The method involves two stages: first sampling $m$ IK solutions for each end-effector pose ($m{=}100$ in our implementation), then applying a dynamic programming algorithm to find joint movements with minimal configurations and joint movements. As shown in Fig \ref{fig: approaches}, we perform IK sampling once and then run the dynamic programming algorithm for each found path. 

\textit{Analysis} --- 
Cart-TSP-IKLink solves the TSP on a graph with $n$ nodes, making it relatively efficient. This approach finds optimal paths in Cartesian space and is ideal for objectives such as minimizing end-effector movements. However, an optimal path in Cartesian space may result in suboptimal paths in joint space. Therefore, Cart-TSP-IKlink may generate motions that involve frequent reconfigurations or large joint movements, which we will demonstrate in \cref{sec:evaluation}.

\vspace{-3mm}
\subsection{Baseline 2: Joint-GTSP} \label{sec:joint_gtsp}
\vspace{-1mm}

Joint-GTSP implements the approach of prior work \cite{hess2012null, kaljaca2020coverage, zhang2024jpmdp}, framing the coverage path planning problem as a Generalized Traveling Salesman Problem (GTSP) in joint space. 
This method consists of three steps: sampling multiple inverse kinematics (IK) solutions for each end-effector target, constructing a graph where nodes represent IK solutions, and finding paths in the graph using a GTSP solver. 

\textit{Step A: Sample IK Solutions} ---
We sample $m$ IK solutions $\{\theta^1_i, \theta^2_i, ...\}$ for each end-effector target $\mathbf{x}_i$ by uniformly sampling initial joint configuration from the robot's joint space. We set $m{=}100$ in our implementation. An optimization-based IK solver, RangedIK \cite{wang2023rangedik}, is used to find IK solutions that are close to the randomly sampled initial configurations. Similar to prior work \cite{wang2024iklink}, we then use DBSCAN \cite{ester1996density} to merge similar IK solutions to reduce the size of the graph.

\textit{Step B: Construct Joint Graph} ---
These IK solutions are used as nodes to construct a graph in joint space. 
All IK solutions for an end-effector target are grouped in a set.
Two IK solutions are connected if their corresponding end-effector targets $\mathbf{x}_i$ and $\mathbf{x}_j$ are neighbors in the input mesh, \textit{i.e.}, $(i,j) {\in} E$. The edge weights are the cost to move between two robot configurations:
    \vspace{-2mm}
\begin{equation}
    w(\mathbf{\theta}_i^a, \mathbf{\theta}_j^b){=} 
    \begin{cases}
        \lVert \mathbf{\theta}_i^a {-} \mathbf{\theta}_j^b \lVert&\text{if } (i.j) {\in} E \text{ and } r(\mathbf{\theta}_i^a, \mathbf{\theta}_j^b ) \\
        M              &\text{if } (i.j) {\in} E \text{ and not }  r(\mathbf{\theta}_i^a,\mathbf{\theta}_j^b ) \\
        \infty & \text{otherwise} 
    \end{cases}
    \vspace{-2mm}
\end{equation}
where $r(\theta^a_i, \theta^b_j)$ decides whether a reconfiguration is required between joint configurations $\theta^a_i$ and $\theta^b_j$. If a reconfiguration is required, a large value $M$ is assigned as the edge weight. 
The function $r(\theta^a, \theta^b)$ returns true if 1) their corresponding end-effector targets are far apart, \textit{i.e.}, $\Delta(\mathbf{x}_i, \mathbf{x}_j) {>} \tau_1$, 2) the maximum joint movement exceeds a threshold, \textit{i.e.,} $\max_k( |\theta_i^a[k] {-} \theta_j^b[k]|) {>} \tau_2$, or 3) the linear interpolation between IK solutions deviates from the end-effector targets, \textit{i.e.},  $\Delta(\text{FK}(\theta), \mathbf{x}_i){>}\tau_3$ or $\Delta(\text{FK}(\theta), \mathbf{x}_j){>} \tau_3$, where $\text{FK}(\cdot)$ is the robot's forward kinematics function, $\mathbf{\theta}{=}\text{lerp}(\theta^a_i, \theta^a_i, t)$ is a linearly interpolated joint configuration and $t {\in} (0,1)$ is the interpolation parameter.   

\textit{Step C: Solve GTSP} --- To find paths that visit each set in the constructed graph exactly once, we introduce an additional node with its own set, connected to all other nodes with zero-weight edges. A GTSP route (a cycle) is converted to a joint path by removing the additional node.

\textit{Analysis} --- The Joint-GTSP approach solves GTSP on a graph with $nm$ nodes, where $n$ is the number of end-effector targets and $m$ is the average number of IK sampling for each end-effector target. Solving an NP-hard problem on a large graph, this approach is slow and does not scale well to large inputs. Many modern GTSP solvers trade solution quality for speed, producing sub-optimal results in shorter timeframes.

\vspace{-1mm}
\section{H-Joint-GTSP} \label{sec:technical_details}
\vspace{-1mm}

Our proposed method, H-Joint-GTSP, reduces the graph size compared to Joint-GTSP by first computing a \textit{guide path} and then sampling inverse kinematics (IK) solutions near it.
As shown in Fig. \ref{fig: approaches}, this approach involves extracting end-effector target exemplars, sampling IK solutions for the exemplars, solving an upper-level GTSP to find guide paths, sampling IK solutions near the guide paths, and solving a lower-level GTSP to find joint motions. 

\textit{Step 1: Extract End-Effector Target Exemplars} ---
We extract key end-effector poses using the affinity propagation clustering algorithm \cite{frey2007clustering} with the distance defined in Equation \ref{eq:cartesian_dist}. The affinity propagation algorithm can automatically determine the number of clusters and return exemplars. The output of the clustering algorithm is $n'$ clusters $\{\mathcal{C}_1, \mathcal{C}_2, ..., \mathcal{C}_{n'} \}$, where the $i$-th cluster involves the indices of $m_i {\geq} 1$ end-effector targets $\mathcal{C}_i {=} \{c_i^1, c_i^2, ..., c_i^{m_i}\}$ and the first item $c_i^1$ corresponds to the center of the cluster.

\textit{Step 2: Sample IK Solutions for Exemplars} ---
Following Step A of Joint-GTSP (\cref{sec:joint_gtsp}), we sample $m$ IK solutions and cluster them for each exemplar (we set $m{=}100$ in our implementation). Additionally, for each IK solution, we verify that the robot can move from the configuration to other end-effector targets in the cluster within certain velocity constraints. Specifically, we use an optimization-based IK solver, RangedIK \cite{wang2023rangedik}, which incorporates velocity constraints and can be expressed as a function $\text{IK}(\mathbf{x}, \theta_{init}, \tau)$. The function returns a valid IK solution that moves the end-effector to the target $\mathbf{x}$ within tolerance $\tau$ and meets the velocity constraint from the initial configuration $\theta_{init}$. If no valid IK solution is found, it returns $\emptyset$. 
We add an IK solution $\theta$ to the set of IK solutions for the $i$-th exemplar $\Theta_i$ only if $\theta$ can reach all targets in the cluster, $\Theta_i {=} \Theta_i \cup \theta$ if $\text{IK}(\mathbf{x}_c, \theta, \tau) {\neq} \emptyset$, $\forall c \in \mathcal{C}_i / c_i^1$.
If $|\Theta_i|{=}0$, we subdivide the cluster $\mathcal{C}_i$ and repeat the sampling process. 

\textit{Step 3: Solve Upper-Level GTSP} ---  
The IK solutions sampled in Step 2 are used as nodes to construct a graph and guide paths are found by solving GTSP, following a process similar to Steps B \& C of Joint-GTSP (\cref{sec:joint_gtsp}). In contrast to Joint-GTSP, which uses IK solutions for all $n$ end-effector targets, here we only use IK samplings for the $n'$ exemplars, resulting in a smaller graph. A guide path $\boldsymbol{\xi}'{=}(\theta'_1, \theta'_2,..., \theta'_{n'}) $ consists of joint configurations that travel through all the exemplars. Given a $\boldsymbol{\xi}'$, we also obtain the order in which the exemplars are visited, denoted by $\boldsymbol{\pi}{=}(\pi_1, \pi_2,..., \pi_{n'})$, where $\pi_i$ is the index of the $i$-th visited exemplar.  Because we use an iteration-based GTSP solver, multiple guide paths are found over time. 

\textit{Step 4: Sample IK Solution near Guide Path} ---
This step generates IK solutions for non-exemplar end-effector targets. Using RangedIK, we greedily propagate to nearby non-exemplars from an IK solution $\theta'_i$  on a guide path. Specifically, we compute $\text{IK}(\mathbf{x}, \theta'_i, \tau)$, where $\mathbf{x}$ represents a non-exemplar end-effector target near $\mathbf{x}_{\pi_i}$. The reachability check in Step 2 ensures that all end-effector targets have at least one IK solution. 

\textit{Step 5: Solve Lower-Level GTSP} ---
In the final setup, we construct a graph and find joint motions by solving the lower-level GTSP. The nodes of this graph are IK solutions on guide paths as well as IK solutions propagated in Step 4. The process follows the same approach as described in Steps B \& C of Joint-GTSP in \cref{sec:joint_gtsp}.

\textit{Analysis} --- The upper-level GTSP graph has $n'm$ nodes, where $n'$ is the number of end-effector target exemplars and $m$ is the average number of sampled IK solutions. The lower-level GTSP graph has $nm'$ nodes, where $n$ is the number of end-effector targets and $m'$ is the number of IK solutions sampled near the guide path ($m'{\geq}1$ because multiple guide paths may be found over time in Step 3, and a guide path may produce multiple IK solutions for an end-effector target in Step 4). Both graphs are smaller than the graph in Joint-GTSP, leading to improved efficiency.

\vspace{-1mm}
\section{Evaluation} \label{sec:evaluation}
\vspace{-1mm}

\begin{figure*} [tb]
  \centering
  \includegraphics[width=7in]{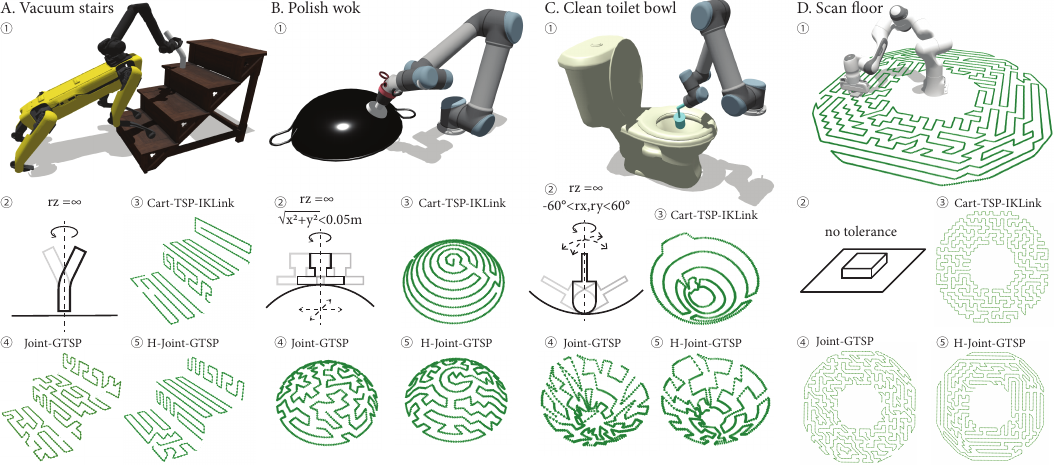}
  \vspace{-5mm}
  \caption{Our experiment involves four benchmark applications. For each, we provide visualizations of \protect\circled{1} the robot performing the task, \protect\circled{2} task tolerances, and \protect\circled{3} \protect\circled{4} \protect\circled{5} how the surface is covered using different approaches. We optimize for the number of reconfigurations (breakpoints on the paths) and joint movements; motions with shorter joint movements tend to result in complex Cartesian space paths.
  The robot visualizations and green traces were generated using \textit{Motion Comparator}\protect\footnotemark \cite{wang2024motion}.
  }
  
  \label{fig: tasks}
  \vspace{-3mm}
\end{figure*}

\begin{table*}[tb]
\caption{Experiment Results and Metrics of the Target Surface} 
\vspace{-3mm}
\label{tab:results}
\begin{center}
\begin{tabular}{c|l|rrr|cc|c}
\hline
Benchmark & \makecell[c]{Method} & \makecell[c]{Mean Num\\ of Reconfig} & \makecell[c]{Mean Joint \\ Movements (rad)$^\dagger$}& \makecell[c]{Mean Comput-\\ation Time (s)} & \makecell{Max Position \\ Error (m)$^\ddagger$} & \makecell{Max Rotation \\ Error (rad)$^\ddagger$} & \makecell{Number of End-\\Effector Targets $n$} \\  
\hline

\rule{0pt}{1.05\normalbaselineskip}%
\multirow{3}{*}{\makecell[c]{Vacuum\\ Stairs}} & Cart-TSP-IKLink & 
    5.00$\pm$0.00 & 32.20$\pm$0.41 \hspace{2mm} & 47.40$\pm$\hspace{1ex}0.35   &  
7.7e-4 & \hspace{1ex}9.5e-3 & 
\multirow{3}{*}{208} \\

& Joint-GTSP & 
5.10$\pm$0.94 & 33.00$\pm$1.11 \hspace{2mm} & 146.22$\pm$35.27   &
 7.9e-4 & 10.0e-3 & \\

& H-Joint-GTSP & 
\textbf{4.30}$\pm$0.46 &  
\textbf{26.66}$\pm$0.51 \hspace{2mm} & 
\textbf{43.59}$\pm$14.07 &
9.8e-4 & \hspace{1ex}9.9e-3 & \\

\hline
\rule{0pt}{1.05\normalbaselineskip}%
\multirow{3}{*}{\makecell[c]{Polish\\Wok}} & Cart-TSP-IKLink &  \textbf{0.00}$\pm$0.00 & 57.78$\pm$0.33 \hspace{2mm} & 99.82$\pm$\hspace{1ex}1.00  &
8.3e-4 & \hspace{1ex}9.2e-3 & 
\multirow{3}{*}{209} \\

& Joint-GTSP & 
14.70$\pm$3.41 & 58.27$\pm$4.93 \hspace{2mm} & 566.26$\pm$95.83  &
7.6e-4 & \hspace{1ex}9.7e-3 & \\

& H-Joint-GTSP & 
\textbf{0.00}$\pm$0.00 & \textbf{41.71}$\pm$0.83 \hspace{2mm} & \textbf{74.56}$\pm$19.71  &
9.6e-4 & 10.0e-3 & \\

\hline
\rule{0pt}{1.05\normalbaselineskip}%
\multirow{3}{*}{\makecell[c]{Clean\\Toilet\\Bowl}} & Cart-TSP-IKLink & 
\textbf{0.00}$\pm$0.00 & 80.78$\pm$1.63 \hspace{2mm} & 89.62$\pm$\hspace{2.2ex}0.37  &
9.0e-4 & n/a & 
\multirow{3}{*}{157} \\

& Joint-GTSP & 
17.29$\pm$6.09 & 61.58$\pm$3.93 \hspace{2mm} & 624.35$\pm$211.82 &
9.5e-4 & n/a & \\

& H-Joint-GTSP & 
\textbf{0.00}$\pm$0.00 & \textbf{17.91}$\pm$1.66 \hspace{2mm} & \textbf{61.35}$\pm$\hspace{2.2ex}9.13  &
9.8e-4 & n/a & \\

\hline
\rule{0pt}{1.05\normalbaselineskip}%
\multirow{3}{*}{\makecell[c]{Scan\\ Floor}} & Cart-TSP-IKLink & 
2.00$\pm$\hspace{1ex}0.00 & 166.23$\pm$0.52 \hspace{2mm} & 262.18$\pm$\hspace{2.2ex}5.12   &
\hspace{1ex}9.9e-4 & \hspace{1ex}9.7e-3 & 
\multirow{3}{*}{674} \\

& Joint-GTSP & 
21.60$\pm$13.71 & 161.29$\pm$9.88 \hspace{2mm} & 1106.65$\pm$198.65   &
\hspace{1ex}9.9e-4 & \hspace{1ex}7.9e-3 & \\

& H-Joint-GTSP & 
\textbf{1.30}$\pm$\hspace{1ex}0.46 & \textbf{123.37}$\pm$1.29 \hspace{2mm} & \textbf{207.84}$\pm$\hspace{1.1ex}69.13  &
10.0e-4 & 10.0e-3 & \\


\hline
\multicolumn{8}{l}{\rule{0pt}{1\normalbaselineskip}%
The range values are standard deviations.
$\dagger$: Joint movements do not involve movements to perform arm reconfigurations.} \\
\multicolumn{8}{l}{
\makecell[l]{$\ddagger$: In our prototype, we set the positional and rotational accuracy of the IK solver as 1e-3 m and 1e-2 rad. As noted in \cref{sec:implementation}, the accuracy can be\\ improved with more computation time. Position and rotation errors are measured only in degrees of freedom without tolerance. }}
\vspace{-3mm}
\end{tabular}
\end{center}
\end{table*}

\begin{figure*} [tb]
  \centering
  \includegraphics[width=6.5in]{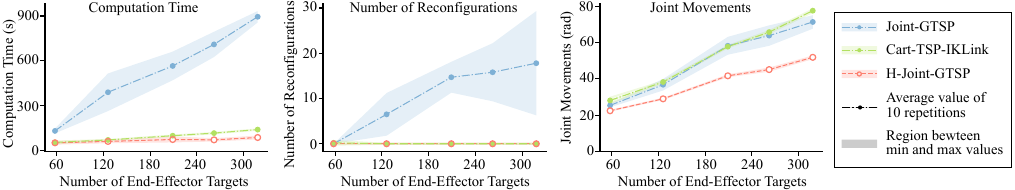}
  \vspace{-1mm}
  \caption{
    To demonstrate the scalability of the approaches, we increased the number of end-effector targets sampled on the surface in the wok polishing task. The results show that the proposed method, H-Joint-GTSP, has superior scalability compared to both Cart-TSP-IKLink and H-Joint-GTSP because it consistently has faster performance and produces higher-quality motion, even when applied to increased sampling density of the input.}
  \label{fig: plot}
  \vspace{-6mm}
\end{figure*}

In this section, we compare our approach, H-Joint-GTSP, with two alternative approaches described in \cref{sec:technical_overview}, Cart-TSP-IKLink and Joint-GTSP, on four simulated benchmark tasks.  

\vspace{-1mm}
\subsection{Implementation Details} \label{sec:implementation}
\vspace{-1mm}

Our prototype system uses the open-source RangedIK\footnote{IK solver: \url{https://github.com/uwgraphics/relaxed\_ik\_core/tree/ranged-ik}} library. RangedIK \cite{wang2023rangedik} is an optimization-based inverse kinematics (IK) solver that can incorporate various requirements such as placing the end-effector within specified task tolerances. RangedIK generates more precise IK solutions with increased iterations. 
In addition, we modified the original GLKH solver \cite{GLKH} to accept a sparse graph as input.
All evaluations were performed on a laptop with Intel Core i9-13950HX 5.50 GHz CPU and 64 GB of RAM.

\vspace{-1mm}
\subsection{Benchmarks}
\vspace{-1mm}
We developed four benchmark applications involving both concave and convex surfaces, as well as 6- and 7-degree-of-freedom (DoF) manipulators, each with varying task tolerances. Fig. \ref{fig: tasks} shows visualizations of these applications. 

\subsubsection{Vacuum Stairs}

This task involves a Boston Dynamics Spot robot vacuuming stairs. The quadruped robot positions its body statically before using its 6 DoF arm to maneuver the vacuum hose. At each stop, the robot cleans both the vertical and horizontal surfaces of two steps. The robot can freely rotate the vacuum hose about its principal axis. 

\subsubsection{Polish Wok} 

This task replicates the wok polishing task in prior work \cite{yang2020cellular, yang2023template}, involving a Universal Robotics UR5 manipulator polishing the outer surface of a wok. In contrast to prior work which locks the robot's last joint to obtain a non-redundant setting, our benchmark has redundancy for coverage path planners to exploit. Specifically, the 6 DoF robot can freely rotate the tool along its principal axis.
In addition, following prior work \cite{schneyer2023segmentation}, we assume that the entire finishing disk, not just the center, can be used for polishing. This enables the robot to move the disk along the tangent plane at the end-effector target (the $xy$ plane) within a specified distance threshold. 

\subsubsection{Clean Toilet Bowl}

This task replicates the toilet bowl cleaning task in prior work \cite{sakata2023coverage} using a 6 DoF Universal Robotics UR5 manipulator.
Due to the semi-spherical shape of the toilet brush head, the robot can freely rotate the brush along its principal axis and adjust its tilt.

\subsubsection{Scan Floor}

This task replicates the floor scanning task in prior work \cite{hyde2023spot} where a robot uses a handheld detector to scan the floor to detect potential chemical or radioactive leaks. 
We use a 7 DoF Franka Emika Panda robot with a rectangular sensor. We assume that the robot needs to precisely control all 6 DoF of the rectangular sensor.

\vspace{-1mm}
\subsection{Experimental Procedure and Results}
\vspace{-1mm}

To account for the randomness in IK sampling and TSP/GTSP solvers, we repeat each benchmark 10 times and report average performance. 
All three approaches use an iteration-based TSP/GTSP solver, \textit{i.e.}, they tend to find improved solutions over time. We consider them converged if motion quality remains unchanged over 30 seconds, at which point we report the motion metrics and computation time.

As shown in Tab. \ref{tab:results}, H-Joint-GTSP consistently outperformed Cart-TSP-IKLink and Joint-GTSP across all four benchmark applications, generating higher-quality motions with shorter computation time.
H-Joint-GTSP produced motions with fewer or equal arm reconfigurations, and when reconfigurations were the same, its motions had shorter joint movements compared to the baseline approaches.
All approaches generated accurate motions.
To further demonstrate the scalability of the approaches, we increased the number of end-effector targets sampled on the surface in the wok polishing task. 
This denser sampling is required when using smaller polishing tools.
As shown in Fig. \ref{fig: plot}, H-Joint-GTSP has better scalability than the baseline approaches.

\vspace{-1mm}
\subsection{Real-Robot Demonstration}
\vspace{-1mm}
To further demonstrate the effectiveness of our proposed approach, we implemented a mock wok polishing task on a physical Universal Robotics UR5 robot. A joint motion $\boldsymbol{\xi}$ found by H-Joint-GTSP was sent to a PID controller to control the robot. As shown in Fig. \ref{fig: teaser} and the supplementary video, the robot successfully covered the bottom of the wok with a smooth and accurate motion.


\vspace{-1mm}
\section{Discussion} \label{sec:discussion}
\vspace{-1mm}
This paper presents a planning approach for effective and efficient joint motion generation for manipulators to cover a surface, aiming to minimize specific joint space costs.

\textit{Limitations} -- Our work has several limitations that suggest potential directions for future research. First, our method uses a heuristic to accelerate the traditional Joint-GTSP approach. While we provide empirical evidence of its efficiency in producing high-quality solutions, we cannot guarantee consistent performance in all scenarios.
Second, our bi-level hierarchical method reduces the size of GTSP. Future research could extend it to multiple levels to further improve performance, though this may produce misleading guide paths.
Third, we observe that both Joint-GTSP and H-Joint-GTSP tend to generate paths with frequent turns, a pattern also observed in the motions of prior work \cite{kaljaca2020coverage, zhang2024jpmdp}.  Future work should explore strategies to balance joint movements with other objectives such as motion smoothness.

\footnotetext{Visualization tool: \url{https://github.com/uwgraphics/MotionComparator}}
\textit{Implications} -- The hierarchical approach presented in this work enables effective and efficient coverage path planning for robot manipulators. 
This approach is beneficial to applications that require dexterous surface coverage, such as sanding, polishing, wiping, and sensor scanning.


\bibliography{root}

\begin{thebibliography}{10}
\providecommand{\url}[1]{#1}
\csname url@samestyle\endcsname
\providecommand{\newblock}{\relax}
\providecommand{\bibinfo}[2]{#2}
\providecommand{\BIBentrySTDinterwordspacing}{\spaceskip=0pt\relax}
\providecommand{\BIBentryALTinterwordstretchfactor}{4}
\providecommand{\BIBentryALTinterwordspacing}{\spaceskip=\fontdimen2\font plus
\BIBentryALTinterwordstretchfactor\fontdimen3\font minus \fontdimen4\font\relax}
\providecommand{\BIBforeignlanguage}[2]{{%
\expandafter\ifx\csname l@#1\endcsname\relax
\typeout{** WARNING: IEEEtran.bst: No hyphenation pattern has been}%
\typeout{** loaded for the language `#1'. Using the pattern for}%
\typeout{** the default language instead.}%
\else
\language=\csname l@#1\endcsname
\fi
#2}}
\providecommand{\BIBdecl}{\relax}
\BIBdecl

\bibitem{hess2012null}
J.~Hess, G.~D. Tipaldi, and W.~Burgard, ``Null space optimization for effective coverage of 3d surfaces using redundant manipulators,'' in \emph{2012 IEEE/RSJ International Conference on Intelligent Robots and Systems}.\hskip 1em plus 0.5em minus 0.4em\relax IEEE, 2012, pp. 1923--1928.

\bibitem{kaljaca2020coverage}
D.~Kaljaca, B.~Vroegindeweij, and E.~Van~Henten, ``Coverage trajectory planning for a bush trimming robot arm,'' \emph{Journal of Field Robotics}, vol.~37, no.~2, pp. 283--308, 2020.

\bibitem{zhang2024jpmdp}
C.~Zhang, H.~Qin, S.~Sun, Y.~Pan, K.~Liu, T.~Li, and X.~Zhao, ``Jpmdp: Joint base placement and multi-configuration path planning for 3d surface disinfection with a uv-c robotic system,'' \emph{Robotics and Autonomous Systems}, vol. 174, p. 104644, 2024.

\bibitem{suarez2018robotsp}
F.~Su{\'a}rez-Ruiz, T.~S. Lembono, and Q.-C. Pham, ``Robotsp--a fast solution to the robotic task sequencing problem,'' in \emph{2018 IEEE International Conference on Robotics and Automation (ICRA)}.\hskip 1em plus 0.5em minus 0.4em\relax IEEE, 2018, pp. 1611--1616.

\bibitem{schneyer2023segmentation}
S.~Schneyer, A.~Sachtler, T.~Eiband, and K.~Nottensteiner, ``Segmentation and coverage planning of freeform geometries for robotic surface finishing,'' \emph{IEEE Robotics and Automation Letters}, vol.~8, no.~8, pp. 5267--5274, 2023.

\bibitem{sakata2023coverage}
Y.~Sakata and T.~Suzuki, ``Coverage motion planning based on 3d model’s curved shape for home cleaning robot,'' \emph{Journal of Robotics and Mechatronics}, vol.~35, no.~1, pp. 30--42, 2023.

\bibitem{moura2018automation}
J.~Moura, W.~Mccoll, G.~Taykaldiranian, T.~Tomiyama, and M.~S. Erden, ``Automation of train cab front cleaning with a robot manipulator,'' \emph{IEEE Robotics and Automation Letters}, vol.~3, no.~4, pp. 3058--3065, 2018.

\bibitem{jing2017sampling}
W.~Jing, J.~Polden, C.~F. Goh, M.~Rajaraman, W.~Lin, and K.~Shimada, ``Sampling-based coverage motion planning for industrial inspection application with redundant robotic system,'' in \emph{2017 IEEE/RSJ International Conference on Intelligent Robots and Systems (IROS)}.\hskip 1em plus 0.5em minus 0.4em\relax IEEE, 2017, pp. 5211--5218.

\bibitem{jing2018computational}
W.~Jing, C.~F. Goh, M.~Rajaraman, F.~Gao, S.~Park, Y.~Liu, and K.~Shimada, ``A computational framework for automatic online path generation of robotic inspection tasks via coverage planning and reinforcement learning,'' \emph{IEEE Access}, vol.~6, pp. 54\,854--54\,864, 2018.

\bibitem{leidner2016robotic}
D.~Leidner, W.~Bejjani, A.~Albu-Sch{\"a}ffer, and M.~Beetz, ``Robotic agents representing, reasoning, and executing wiping tasks for daily household chores,'' \emph{AUTONOMOUS AGENTS AND MULTI-AGENT SYSTEMS}, 2016.

\bibitem{thakar2022area}
S.~Thakar, R.~K. Malhan, P.~M. Bhatt, and S.~K. Gupta, ``Area-coverage planning for spray-based surface disinfection with a mobile manipulator,'' \emph{Robotics and Autonomous Systems}, vol. 147, p. 103920, 2022.

\bibitem{do2023geometry}
V.-T. Do and Q.-C. Pham, ``Geometry-aware coverage path planning for depowdering on complex 3d surfaces,'' \emph{IEEE Robotics and Automation Letters}, vol.~8, no.~9, pp. 5552--5559, 2023.

\bibitem{yang2024improved}
T.~Yang, J.~V. Miro, Y.~Wang, and R.~Xiong, ``An improved maximal continuity graph solver for non-redundant manipulator non-revisiting coverage,'' \emph{IEEE Trans. on Automation Science and Engineering}, 2024.

\bibitem{yang2020cellular}
T.~Yang, J.~V. Miro, Q.~Lai, Y.~Wang, and R.~Xiong, ``Cellular decomposition for nonrepetitive coverage task with minimum discontinuities,'' \emph{IEEE/ASME Transactions on Mechatronics}, vol.~25, no.~4, pp. 1698--1708, 2020.

\bibitem{yang2020non}
T.~Yang, J.~V. Miro, Y.~Wang, and R.~Xiong, ``Non-revisiting coverage task with minimal discontinuities for non-redundant manipulators,'' in \emph{16th Conf. on Robotics-Science and Systems}.\hskip 1em plus 0.5em minus 0.4em\relax MIT PRESS, 2020.

\bibitem{saha2006planning}
M.~Saha, T.~Roughgarden, J.-C. Latombe, and G.~S{\'a}nchez-Ante, ``Planning tours of robotic arms among partitioned goals,'' \emph{The International Journal of Robotics Research}, vol.~25, no.~3, pp. 207--223, 2006.

\bibitem{alatartsev2015robotic}
S.~Alatartsev, S.~Stellmacher, and F.~Ortmeier, ``Robotic task sequencing problem: A survey,'' \emph{Journal of intelligent \& robotic systems}, vol.~80, pp. 279--298, 2015.

\bibitem{song2018surface}
S.~Song and S.~Jo, ``Surface-based exploration for autonomous 3d modeling,'' in \emph{2018 IEEE international conference on robotics and automation (ICRA)}.\hskip 1em plus 0.5em minus 0.4em\relax IEEE, 2018, pp. 4319--4326.

\bibitem{obermeyer2012sampling}
K.~J. Obermeyer, P.~Oberlin, and S.~Darbha, ``Sampling-based path planning for a visual reconnaissance unmanned air vehicle,'' \emph{Journal of Guidance, Control, and Dynamics}, vol.~35, no.~2, pp. 619--631, 2012.

\bibitem{dhami2024gatsbi}
H.~Dhami, C.~Reddy, V.~D. Sharma, T.~Williams, and P.~Tokekar, ``Gatsbi: An online gtsp-based algorithm for targeted surface bridge inspection and defect detection,'' \emph{arXiv preprint arXiv:2406.16625}, 2024.

\bibitem{bahnemann2021revisiting}
R.~B{\"a}hnemann, N.~Lawrance, J.~J. Chung, M.~Pantic, R.~Siegwart, and J.~Nieto, ``Revisiting boustrophedon coverage path planning as a generalized traveling salesman problem,'' in \emph{The 12th Conference on Field and Service Robotics}.\hskip 1em plus 0.5em minus 0.4em\relax Springer, 2021, pp. 277--290.

\bibitem{cao2020hierarchical}
C.~Cao, J.~Zhang, M.~Travers, and H.~Choset, ``Hierarchical coverage path planning in complex 3d environments,'' in \emph{IEEE Int. Conf. on Robotics and Automation (ICRA)}.\hskip 1em plus 0.5em minus 0.4em\relax IEEE, 2020, pp. 3206--3212.

\bibitem{pop2023comprehensive}
P.~C. Pop, O.~Cosma, C.~Sabo, and C.~P. Sitar, ``A comprehensive survey on the generalized traveling salesman problem,'' \emph{European Journal of Operational Research}, 2023.

\bibitem{pop2007new}
P.~C. Pop, ``New integer programming formulations of the generalized traveling salesman problem,'' \emph{American Journal of Applied Sciences}, vol.~4, no.~11, pp. 932--937, 2007.

\bibitem{noon1991lagrangian}
C.~E. Noon and J.~C. Bean, ``A lagrangian based approach for the asymmetric generalized traveling salesman problem,'' \emph{Operations Research}, vol.~39, no.~4, pp. 623--632, 1991.

\bibitem{kara2012new}
I.~Kara, H.~Guden, and O.~N. Koc, ``New formulations for the generalized traveling salesman problem,'' in \emph{Proceedings of the 6th international conference on applied mathematics, simulation, modelling, ASM}, vol.~12, 2012, pp. 60--65.

\bibitem{hu2008effective}
B.~Hu and G.~R. Raidl, ``Effective neighborhood structures for the generalized traveling salesman problem,'' in \emph{Evolutionary Computation in Combinatorial Optimization: 8th European Conference, Naples, Italy, March 26-28, 2008. Proceedings 8}.\hskip 1em plus 0.5em minus 0.4em\relax Springer, 2008, pp. 36--47.

\bibitem{helsgaun2015solving}
K.~Helsgaun, ``Solving the equality generalized traveling salesman problem using the lin--kernighan--helsgaun algorithm,'' \emph{Mathematical Programming Computation}, vol.~7, pp. 269--287, 2015.

\bibitem{snyder2006random}
L.~V. Snyder and M.~S. Daskin, ``A random-key genetic algorithm for the generalized traveling salesman problem,'' \emph{European journal of operational research}, vol. 174, no.~1, pp. 38--53, 2006.

\bibitem{GLKH}
\BIBentryALTinterwordspacing
K.~Helsgaun. (2021) Glkh-a program for solving the equality generalized traveling salesman problem (e-gtsp). [Online]. Available: \url{http://webhotel4.ruc.dk/~keld/research/GLKH/}
\BIBentrySTDinterwordspacing

\bibitem{smith2017glns}
S.~L. Smith and F.~Imeson, ``Glns: An effective large neighborhood search heuristic for the generalized traveling salesman problem,'' \emph{Computers \& Operations Research}, vol.~87, pp. 1--19, 2017.

\bibitem{sadeghian2013task}
H.~Sadeghian, L.~Villani, M.~Keshmiri, and B.~Siciliano, ``Task-space control of robot manipulators with null-space compliance,'' \emph{IEEE Transactions on Robotics}, vol.~30, no.~2, pp. 493--506, 2013.

\bibitem{claret2017exploiting}
J.-A. Claret, G.~Venture, and L.~Basa{\~n}ez, ``Exploiting the robot kinematic redundancy for emotion conveyance to humans as a lower priority task,'' \emph{International journal of social robotics}, vol.~9, no.~2, pp. 277--292, 2017.

\bibitem{wang2025anytime}
Y.~Wang and M.~Gleicher, ``Anytime planning for end-effector trajectory tracking,'' \emph{IEEE Robotics and Automation Letters}, vol.~10, no.~4, pp. 3246--3253, 2025.

\bibitem{wang2023exploiting}
Y.~Wang, C.~Sifferman, and M.~Gleicher, ``Exploiting task tolerances in mimicry-based telemanipulation,'' in \emph{IEEE/RSJ International Conf. on Intelligent Robots and Systems (IROS)}.\hskip 1em plus 0.5em minus 0.4em\relax IEEE, 2023, pp. 7012--7019.

\bibitem{wang2024iklink}
Y.~Wang, C.~Sifferman, and G.~Michael, ``Iklink: End-effector trajectory tracking with minimal reconfigurations,'' in \emph{2024 IEEE International Conference on Robotics and Automation (ICRA)}.\hskip 1em plus 0.5em minus 0.4em\relax IEEE, 2024.

\bibitem{helsgaun2000effective}
K.~Helsgaun, ``An effective implementation of the lin--kernighan traveling salesman heuristic,'' \emph{European journal of operational research}, vol. 126, no.~1, pp. 106--130, 2000.

\bibitem{wang2023rangedik}
Y.~Wang, P.~Praveena, D.~Rakita, and M.~Gleicher, ``Rangedik: An optimization-based robot motion generation method for ranged-goal tasks,'' in \emph{2023 IEEE International Conference on Robotics and Automation (ICRA)}.\hskip 1em plus 0.5em minus 0.4em\relax IEEE, 2023, pp. 8090--8096.

\bibitem{ester1996density}
M.~Ester, H.-P. Kriegel, J.~Sander, X.~Xu \emph{et~al.}, ``A density-based algorithm for discovering clusters in large spatial databases with noise,'' in \emph{kdd}, vol.~96, no.~34, 1996, pp. 226--231.

\bibitem{frey2007clustering}
B.~J. Frey and D.~Dueck, ``Clustering by passing messages between data points,'' \emph{science}, vol. 315, no. 5814, pp. 972--976, 2007.

\bibitem{wang2024motion}
Y.~Wang, A.~Peseckis, Z.~Jiang, and M.~Gleicher, ``Motion comparator: Visual comparison of robot motions,'' \emph{IEEE Robotics and Automation Letters}, vol.~9, no.~9, pp. 7699--7706, 2024.

\bibitem{yang2023template}
T.~Yang, J.~V. Miro, M.~Nguyen, Y.~Wang, and R.~Xiong, ``Template-free nonrevisiting uniform coverage path planning on curved surfaces,'' \emph{IEEE/ASME Transactions on Mechatronics}, vol.~28, no.~4, pp. 1853--1861, 2023.

\bibitem{hyde2023spot}
J.~Hyde, J.~Bloodwood, C.~Abeyta, and M.~Dierauer, ``Spot robot staffing augmentation at los alamos national laboratory,'' LA-UR-23-24724, Los Alamos National Laboratory.\hskip 1em plus 0.5em minus 0.4em\relax Office of Scientific and Technical Information (OSTI), 2023.

\end{thebibliography}
\bibliographystyle{IEEEtran}

\end{document}